\documentclass[conference]{IEEEtran}
\IEEEoverridecommandlockouts
\usepackage{cite}
\usepackage{amsmath,amssymb,amsfonts}
\usepackage{algorithmic}
\usepackage{graphicx}
\usepackage{textcomp}
\usepackage{xcolor}
\usepackage{hyperref}

\def\BibTeX{{\rm B\kern-.05em{\sc i\kern-.025em b}\kern-.08em
    T\kern-.1667em\lower.7ex\hbox{E}\kern-.125emX}}
\begin{document}

\title{Green AI: Exploring Carbon Footprints, Mitigation Strategies, and Trade Offs in Large Language Model Training\\
{\footnotesize \textsuperscript{*}Note: Sub-titles are not captured in Xplore and
should not be used}
}

\author{
\IEEEauthorblockN{Vivian Liu}
\IEEEauthorblockA{Independent Researcher \\
vivianliu292@gmail.com}
\and
\IEEEauthorblockN{Yiqiao Yin}
\IEEEauthorblockA{Columbia University \\
University of Chicago \\
New York, US \\
yy2502@columbia.edu}
}

\maketitle

\begin{abstract}
Prominent works in the field of Natural Language Processing have long attempted to create new innovative models by improving upon previous model training approaches, altering model architecture, and developing more in-depth datasets to better their performance. However, with the quickly advancing field of NLP comes increased greenhouse gas emissions, posing concerns over the environmental damage caused by training LLMs. Gaining a comprehensive understanding of the various costs, particularly those pertaining to environmental aspects, that are associated with artificial intelligence serves as the foundational basis for ensuring safe AI models. Currently, investigations into the CO2 emissions of AI models remain an emerging area of research, and as such, in this paper, we evaluate the CO2 emissions of well-known large language models, which have an especially high carbon footprint due to their significant amount of model parameters. We argue for the training of LLMs in a way that is responsible and sustainable by suggesting measures for reducing carbon emissions. Furthermore, we discuss how the choice of hardware affects CO2 emissions by contrasting the CO2 emissions during model training for two widely used GPUs. Based on our results, we present the benefits and drawbacks of our proposed solutions and make the argument for the possibility of training more environmentally safe AI models without sacrificing their robustness and performance. 
\end{abstract}


\section{Introduction}
From language translation to smart assistants, the application of large language models (LLMs) in our daily lives has greatly increased. In recent years, the development of LLMs is rapidly expanding, with more and more large corporations hopping on the train of fine-tuning natural language processing (NLP) models. As such, making existing models more efficient and optimizing the performance of models has been a major area of focus \cite{brown2020language}. Recent works have looked into improving upon previous methods of training models \cite{liu2019roberta}. For example, a new pretraining approach uses replaced token detection and corrupts the model input by replacing some tokens with plausible alternatives sampled from a small generator network, leading to contextual representations that substantially outperform ones learned by other models given the same model size, data, and compute \cite{clark2020electra}. Other approaches for improving performance include utilizing new datasets, like WebText, which uses millions of webpages to teach language models natural language processing tasks without explicit supervision \cite{radford2019language}. Research emphasizing techniques involving generative pre-training of a language model on a diverse corpus of unlabeled text, followed by discriminative fine-tuning has also shown results in commonsense reasoning, question answering, and textual entailment as well \cite{radford2018improving}. The research done to make these advancements in the NLP field aim mainly to improve upon NLP models or the way they are trained, but only through a lens of achieving the best performance possible. However, many high-performance NLP models also have a staggeringly large amount of model parameters, and this alludes to an expanding concern for LLMs: how \textbf{environmentally safe} training these models is.

Model training for large language models (such as attention-based NLP models) is highly costly and it opens up a range of ethical issues \cite{vaswani2017attention}. As the NLP field expands, leading to increasingly impressive breakthroughs and higher-performing models, so do the costs of the extensive training involved in training these models. With our increased reliance on LLMs, the CO2 emissions caused by training NLP models are an issue that absolutely must be discussed in order to drive forward the development of safe AI.

Some large language models have billions of parameters yet we functionally treat them as black boxes \cite{huang2023chatgpt}. When using these models, we focus on the results achieved without questioning the various costs of training the models, causing the impact they have on our environment to remain largely undiscussed. In regards to these issues, this paper proposes an experiment using Code Carbon's CO2 emissions tracker to evaluate the CO2 emissions of models as they train and investigate the pros and cons of fine-tuning LLMs \cite{budennyy2022eco2ai}. Further, our work assesses the performance of the models with two different data sets and scores them using cosine similarity, a measure of similarity between vectors, and semantic textual similarity (STS), a measure of semantic equivalence between two texts \cite{xia2015learning, agirre2012semeval}.

Aside from carbon emissions, other considerations we account for include the financial implications of implementing each strategy to lower emissions. Challenges with utilizing transfer learning also arise in effectively deploying LLMs under resource limitations, particularly when operating within constrained computational training and inference budgets \cite{sanh2019distilbert}. As such, our objective is to discuss these aspects of possible solutions to evaluate the feasibility of applying them to lower CO2 emissions. Moreover, we seek to ascertain whether our proposed strategies for mitigating environmental damage are truly accessible to the layperson aspiring to train LLMs.

Based on our experiment, we are able to present the analysis of different strategies to lower CO2 emissions. The results from the training and testing of LLMs lead us to propose that balancing impeccably robust and high-performing models with strategies that commendably reduce CO2 emissions to ensure a sustainable future is not just a mere possibility, but a tangible reality within our reach.

\section{Related Work}
Previous research has attempted to make models lighter and more efficient. For example, Albert is a lighter version of the well-known BERT model with reduced parameters and faster training times \cite{lan2019albert}. DistilBERT similarly reduces the parameters of the BERT model, lowering model parameters by 40\% \cite{sanh2019distilbert}. Using these lighter models may help us reduce the carbon footprint of model training, and as such these papers are undeniably taking a step in the right direction by creating models that are lighter and more affordable to train. However, much of this research focuses on lowering the monetary cost of training and for the most part neglects to discuss the importance of decreasing CO2 emissions.

Aside from that, there has been research done to create carbon emissions trackers and calculators that can approximate the CO2 emissions of models \cite{lacoste2019quantifying}. This has led to the release of open-source packages intended to help data scientists and researchers track the carbon emissions of their models \cite{budennyy2022eco2ai}. Other recent works have further attempted to start shedding light on the importance of lowering the carbon emissions produced in model training by surveying factors that may influence the CO2 emissions produced by machine learning models \cite{luccioni2023counting}.

\subsection{Motivation}
Apprehensions arise from the rapid advancements witnessed in the NLP domain. In the absence of timely attention to environmental concerns, there exists a legitimate concern that this unchecked progress may result in irreversible environmental harm in the foreseeable future. Thus, our goal is to generate usable metrics quantifying model performance and CO2 emissions which will then enable corporations to utilize large language models to make informed and sustainable decisions when training NLP models. We also attempt to analyze different approaches for reducing CO2 emissions to help individuals take actionable efforts to lower the CO2 emissions when training their NLP models.

We additionally attempt to present an accurate picture of the pricing of possible solutions for lowering CO2 emissions. By analyzing the financial costs of implementing certain strategies to lower CO2 emissions, we can answer the question, "Is training AI sustainably truly something that anybody can do, or is it only a hope for large corporations which have better access to more novel solutions for increasing CO2 emission efficiency?"

\section{Experimental Setup}

To formulate strategies for reducing the CO2 emissions of model training and determine whether they are financially accessible to any person wanting to train LLMs, an initial prerequisite involves the selection of suitable models for performance and emissions assessment. Furthermore, the selection of appropriate datasets for model training and subsequent evaluation, as well as the designation of a set of quantifiable metrics to gauge model proficiency, are essential determinations in this process.

\begin{figure}[ht]
    \centering
    \includegraphics[width=1\linewidth]{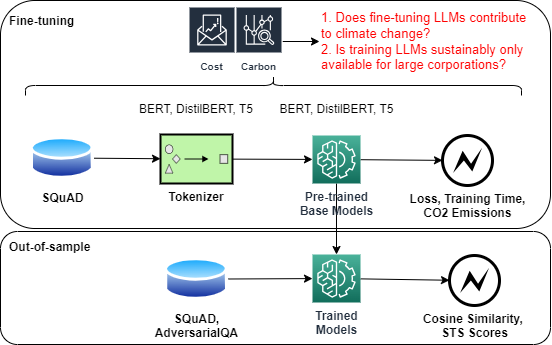}
    \caption{Flowchart of the process used for training, testing, and recording measurements of model performance.}
    \label{fig:main-graph}
\end{figure}

The principal stages of the experiment are depicted in Figure \ref{fig:main-graph}. This experiment involves the fine-tuning of a variety of large language models, including BERT, DistilBERT, and T5, using different tokenizers and the SQuAD dataset. Throughout the fine-tuning phase, parameters such as loss, training time, and CO2 emissions are meticulously documented. Subsequently, the performance of the trained models is assessed using an out-of-sample test set derived from both the SQuAD and AdversarialQA datasets. Evaluation metrics employed for this purpose include cosine similarity and Semantic Textual Similarity (STS) scores.

\subsection{Models}
Transformer models combined with self-supervised pretraining have revolutionized natural language processing (NLP) and information retrieval (IR) \cite{lin2022pretrained}. They produce high-quality results across various domains and tasks. Due to this, we selected three well-known transformers to evaluate the performance of: BERT, DistilBERT, and T5.

BERT, short for Bidirectional Encoder Representations from Transformers, is a model pre-trained on Masked Language Modeling and Next Sentence Prediction, enabling it to excel in various NLP tasks. BERT is designed to pre-train deep bidirectional representations from unlabeled texts by conditioning on both the left and right context in all layers. As a result, the pre-trained BERT model can be fine-tuned with just one additional output layer to create state-of-the-art models for a wide range of tasks, such as question answering and language inference, without substantial task-specific architecture modifications. BERT has obtained new state-of-the-art results on eleven natural language processing tasks, including achieving improvements on previous models in areas such as GLUE score, MultiNLI accuracy, SQUAD v1.1 question answering Test F1, and SQUAD v2.0 Test F1 \cite{devlin2018bert}. BERT’s attention heads exhibit patterns such as attending to delimiter tokens, specific positional offsets, or broadly attending over the whole sentence. Certain attention heads correspond well to linguistic notions of syntax and coreference \cite{clark2019does}.

DistilBERT, on the other hand, represents a distilled version of the BERT model, employing knowledge distillation to significantly reduce the number of model parameters and create a more lightweight version. Distillation is a method to pre-train a smaller general-purpose language representation model called DistilBERT, which can then be fine-tuned with good performances on a wide range of tasks like its larger counterparts. The DistilBERT model is created using knowledge distillation and is leveraged during the pre-training phase to reduce the size of a BERT model by 40\%, while retaining 97\% of its language understanding capabilities and being 60\% faster. A triple loss combining language modeling, distillation, and cosine-distance losses is introduced to leverage the inductive biases learned by larger models during pre-training. The smaller, faster, and lighter model is cheaper to pre-train \cite{sanh2019distilbert}.

T5, the fifth generation of a transformer model, is capable of performing tasks presented to it in a text-to-text format \cite{bird2023chatbot}. To further explore the models' performance, we also conducted experiments with different tokenizers: bert-base-cased, distilbert-base-uncased, and t5-base tokenizers. These tokenizer variations allowed us to investigate how tokenization strategies might impact the models' effectiveness in handling text data and CO2 emissions.

Beyond the base models themselves, we also tried using BERT's tokenizer with a DistilBERT base model and we combined DistilBERT's tokenizer with a BERT base model. This was done to develop an understanding of how the tokenizer selected affects a model's CO2 emissions and model performance.

\subsection{Data}
The pre-trained models selected were trained using the Stanford Question Answering Dataset (SQUAD), a dataset comprising of over 100,000 questions created by crowdworkers that is based on Wikipedia articles \cite{rajpurkar2016squad}. Each question from the dataset requires an answer that can be found within the corresponding reading passage in order to test reading comprehension. The dataset is analyzed to determine the reasoning types needed to answer the questions, with a particular focus on dependency and constituency trees. Under testing, it was found that human performance on the dataset greatly outperformed the performance of a robust logistic regression model and a baseline model, indicating that the dataset serves as a challenging problem for future research in the field of reading comprehension \cite{rajpurkar2016squad}. 

Before training the models, the data was split into training and validation, with a training-validation split of 20\% used for model training. This meant that of the 107,785 question-answer pairs contained in the SQUAD dataset, 21,557 question-answer pairs were used for evaluating the models' performance between epochs, while the remaining 86,228 question-answer pairs were used purely for the actual training of each model.

Furthermore, we utilized the AdversarialQA dataset as a held-out test set to gauge the performance of our trained NLP models on data to which they had not been previously exposed. The AdversarialQA dataset was conceived through an adversarial human annotation process, which incorporated both a human annotator and an NLP model in the creation of question-answer pairs. Designed to scrutinize a model's reading comprehension capabilities, this dataset comprises approximately 36,000 question-answer pairs \cite{bartolo2020beat}.

\subsection{Metrics}
As previously mentioned, to gain a comprehensive understanding of the performance of various models and to compare their performance with their corresponding carbon emissions, we employed the emissions tracker from the CodeCarbon Python package to provide us with an estimate for the amount of CO2 emissions generated during model training. From the training of the various models, we recorded essential metrics such as loss, training time, CO2 emissions, RAM usage, and model parameters (presented in Table \ref{tab:model-a-tokenizer-b-test-set-performance-T4} and Table \ref{tab:model-a-tokenizer-b-test-set-performance-A100}).

The first table provides an overview of the models' performance when trained using the T4 GPU (see detailed results in Table \ref{tab:model-a-tokenizer-b-test-set-performance-T4}). It includes the validation loss measured for each model and the corresponding estimated amount of CO2 emissions in kilograms generated during the training process.

In contrast, the second table showcases the results of training the models with the A100 40GB SXM GPU (see detailed results in Table \ref{tab:model-a-tokenizer-b-test-set-performance-A100}). It also presents the validation loss and the estimated CO2 emissions produced during the training sessions. By juxtaposing these two tables, we can gain valuable insights into the trade-offs between model performance and environmental impact based on the type of GPU utilized during the training process.

In the third table, we conducted 100 random samplings from the SQUAD dataset to evaluate the performance of the different models (see detailed results in Table \ref{tab:test-models-on-squad-data}). Each model was tasked with making inferences using 100 queries from the dataset, and we measured their performance using cosine similarity, STS (BERT-Embedding), STS (OpenAI-Embedding), and STS (Palm-Embedding). The formula for cosine similarity is defined in the following equation. \begin{equation} similarity=\cos(\theta) = \frac{\mathbf{A} \cdot \mathbf{B}}{||\mathbf{A}|| \times ||\mathbf{B}||} \end{equation} The values presented in the table represent the average performance scores, and the standard deviation is listed in parentheses. These results were obtained by calculating the models' performance over 100 samples, providing a comprehensive assessment of their overall effectiveness and consistency in handling the dataset queries.

In the fourth table, we continue the evaluation of model performance over 100 samples (see detailed results in Table \ref{tab:test-models-on-adversarialQA-data}). However, in this case, the models were assessed using the AdversarialQA dataset, which serves as a held-out test set, rather than the SQUAD dataset. Each model was subjected to the same inference process, and their performance was measured using cosine similarity, STS (BERT-Embedding), STS (OpenAI-Embedding), and STS (Palm-Embedding) metrics. Similar to the third table, the values in the fourth table represent the average performance scores, with the standard deviation indicated in parentheses. This evaluation using the AdversarialQA dataset allows us to further validate the models' capabilities in handling different types of queries and assess their robustness across diverse datasets.

\section{Results and Discussion}

We use the measurements recorded from model training, testing, and the CO2 emissions tracker to weigh the pros and cons of certain models and GPUs. By analyzing the benefits of some models, conclusions can be formulated about what methods are effective for lowering CO2 emissions.

\subsection{Climate and Environmental Protection}

Within the scope of our experiment, we discuss the environmental consequences of model training in the form of CO2 emissions and use the CO2 emissions estimates provided by Code Carbon's emissions tracker to quantify how sustainable certain models and GPU options are.

\begin{table}
    \centering
    \resizebox{1\columnwidth}{!}{   
        \begin{tabular}{c|c|c|c|c|c|c}
        \hline
          Model	&	Tokenizer	&	Loss	&	Time	&	CO2 (kg)	&	RAM (W)	&	Model Param.	\\	\hline
            B.	&	D.	&	1.5	&	1152	&	1.15E-02	&	4.8	&	1.1 bn	\\	\hline
            B.	&	B.	&	3.0	&	1183	&	1.18E-02	&	9.5	&	1.1 bn	\\	\hline
            D.	&	D.	&	1.4	&	634	&	6.28E-03	&	9.5	&	66 mm	\\	\hline
            D.	&	B.	&	3.3	&	646	&	1.12E-02	&	9.5	&	66 mm	\\	\hline
            t5	&	t5-base	&	1.0	&	22996	&	8.07E-02	&	9.5	&	11 bn	\\	\hline
        \end{tabular}
    }
    
    \caption{\textbf Model performance trained on SQUAD data, 3 epochs, and T4 GPU. Model consists of BERT (named as B.), DistilBERT (named as D.), and T5. Tokenizer consists of bert-base-cased (named as B.), distilbert-base-uncased (named as D.), and t5-base. Loss is the validation loss and time is the time taken in seconds for training. CO2 is CO2 emissions and RAM is RAM usage from training. Model Param. is the model parameters of each model. The name ``bn'' and ``mm'' represent billion and million, respectively. }
    \label{tab:model-a-tokenizer-b-test-set-performance-T4} 
\end{table}

\begin{table}
    \centering
    \resizebox{1\columnwidth}{!}{   
        \begin{tabular}{c|c|c|c|c|c|c}
        \hline
           Model	&	Tokenizer	&	Loss	&	Time	&	CO2 (kg)	&	RAM (W)	& Param.	\\	\hline
            B.	&	D.	&	1.4	&	221	&	3.20E-05	&	31.3	&	1.1 bn	\\	\hline
            B.	&	B.	&	3.1	&	233	&	4.40E-03	&	31.3	&	1.1 bn	\\	\hline
            D.	&	D.	&	1.5	&	142	&	1.60E-05	&	31.3	&	66 mm	\\	\hline
            D.	&	B.	&	3.2	&	598	&	5.30E-03	&	9.5	&	66 mm	\\	\hline
            t5	&	t5-base	&	1.0	&	7661	&	3.70E-04	&	31.3	&	11 bn	\\	\hline
        \end{tabular}
    }
    
    \caption{\textbf Model performance trained on SQUAD data, 3 epochs, and A100 GPU. Model consists of BERT (named as B.), DistilBERT (named as D.), and T5. Tokenizer consists of bert-base-cased  (named as B.), distilbert-base-uncased (named as D.), and t5-base. Loss is the validation loss and time is the time taken in seconds for training. CO2 is CO2 emissions and RAM is RAM usage from training. Param. is the model parameters of each model. The name ``bn'' and ``mm'' represent billion and million, respectively.}
    \label{tab:model-a-tokenizer-b-test-set-performance-A100} 
\end{table}

The experimental results reveal that both the T5 and BERT models emitted considerably more CO2 compared to DistilBERT (see Table \ref{tab:model-a-tokenizer-b-test-set-performance-T4} and Table \ref{tab:model-a-tokenizer-b-test-set-performance-A100}). Similarly, utilizing a bert-base-cased tokenizer also resulted in higher CO2 emissions compared to using a distilbert-base-uncased tokenizer. During model training with the T4 GPU, the DistilBERT model with a distilbert-base-uncased tokenizer had 46.9\% fewer CO2 emissions in comparison to the BERT model with a tokenizer of bert-base-cased.

Contrasting tables\ref{tab:model-a-tokenizer-b-test-set-performance-T4} and \ref{tab:model-a-tokenizer-b-test-set-performance-A100} clearly demonstrates the substantial reduction in model training time and CO2 emissions when switching from the T4 GPU to the A100 GPU. On average the A100 GPU decreased model training time by 62.6\%. It produced especially large reductions in training time for the BERT model with a tokenizer of distilbert-base-uncased and the BERT model with a tokenizer of bert-base-cased which experienced 80.8\% and 80.3\% reductions in training time respectively.

The A100 GPU also lowered carbon emissions by a staggering 83\% on average across all five models. It remarkably reduced CO2 emissions by 99.7\% for both BERT with a tokenizer of distilbert-base-uncased and DistilBERT with a tokenizer of distilbert-base-uncased, 67.2\% for BERT with a tokenizer of bert-base-cased, 53.1\% for DistilBERT with a tokenizer of bert-base-cased, and 99.5\% for the T4 model. 

To contextualize the effects of the CO2 emissions of each model, we can look at the CO2 emissions of a human. For a year, a human exhales about 255 kg of CO2. Comparing this to our results, our experiment demonstrates that BERT with a tokenizer of distilbert-base-uncased produces 0.0115 kg of CO2 by training on 3 epochs with a T4 GPU. Based off this, for corporations to train large language models with 1000 epochs, model training would produce about 3.83 kg of CO2. In 2022, there were 10 LLMs released, which would be 3.83*10 kg of CO2 emissions if each model was trained for 1000 epochs, which is 38.3 kg for 10 models trained with 1000 epochs each. If a large corporation has a whole department of 1000 people all training their own LLMs, this would total 33300 kg per year for one corporation.





\subsection{Performance}

Apart from CO2 emissions, another key consideration is the performance of models and the extent to which we can lower CO2 emissions while preserving high performance. Despite being more lightweight compared to its BERT counterpart, models using a distilbert-base-uncased tokenizer demonstrated better performance than models using a bert-base-cased tokenizer on the SQUAD and AdversarialQA datasets. Models with a distilbert-base-uncased tokenizer also exhibited lower loss and higher cosine similarity, STS (BERT-Embedding), STS (OpenAI-Embedding), and STS (Palm-Embedding) scores. Beyond decreasing carbon emissions, the DistilBERT model with a distilbert-base-uncased tokenizer lowered the time taken to train by 46\% and decreased loss by 54.5\% compared to the BERT model with bert-base-cased as its tokenizer. This model produced similar reductions in training time, carbon emissions, and loss compared to BERT with a bert-base-cased tokenizer when trained while using the A100 GPU. The BERT model with a tokenizer of distilbert-base-uncased also had lower loss than the BERT model with a bert-base-cased tokenizer, experiencing a 51.3\% decrease in loss. These consistent findings underscore that the distilbert-base-uncased tokenizer did not sacrifice any of the efficacy of the bert-base-cased tokenizer, as it consistently outperforms models employing the bert-base-cased tokenizer across various evaluation metrics and datasets.

\begin{table}[ht]
    \centering
    \resizebox{1\columnwidth}{!}{   
        \begin{tabular}{c|c|c|c|c|c|c}
        \hline
          Model	&	Token.	&	Param.	&	Cosine	&	BERT	&	OpenAI	&	Palm	\\	\hline
            B.	&	D.	&	1.1 bn	&	0.27 (0.39)	&	0.38 (0.38)	&	0.82 (0.10)	&	0.67 (0.17)	\\	\hline
            B.	&	B.	&	1.1 bn	&	0.03 (0.14)	&	0.20 (0.19)	&	0.78 (0.05)	&	0.59 (0.09)	\\	\hline
            D.	&	D.	&	66 mm	&	0.20 (0.30)	&	0.34 (0.31)	&	0.81 (0.08)	&	0.65 (0.14)	\\	\hline
            t5	&	t5-base	&	11 bn	&	0.23 (0.26)	&	0.54 (0.25)	&	0.88 (0.06)	&	0.77 (0.10)	\\	\hline
        \end{tabular}
    }
    
    \caption{\textbf Model performance scored on 100 observations sampled randomly from SQUAD data. Model consists of BERT (named as B.), DistilBERT (named as D.), and T5. Tokenizer (named as token.) consists of bert-base-cased (named as B.), distilbert-base-uncased (named as D.), and t5-base. Param. is the model parameters of each model. Cosine is the cosine similarity, BERT is the STS (BERT) score, OpenAI is the STS (OpenAI) score, and Palm is the STS (Palm) score. The name ``bn'' and ``mm'' refer to billion and million, respectively.}
    \label{tab:test-models-on-squad-data}
\end{table}

By training the T5 model on the SQUAD dataset for three epochs, it was able to perform better in validation loss than the other models trained on three epochs. However, in Table \ref{tab:test-models-on-squad-data} and Table \ref{tab:test-models-on-adversarialQA-data}, the T5 model trained for only one epoch does not perform the best in all scoring categories. The experimental results reveal the general correlation between the number of epochs a model is trained for and its performance. Unfortunately, training for more epochs subsequently leads to higher CO2 emissions.

\begin{table}[ht]
    \centering
    \resizebox{1\columnwidth}{!}{   
        \begin{tabular}{c|c|c|c|c|c|c}
        \hline
           Model	&	Token.	& Param.	&	Cosine	&	BERT	&	OpenAI	&	Palm	\\	\hline
            B.	&	D.	&	1.1 bn	&	0.11 (0.23)	&	0.28 (0.26)	&	0.80 (0.06)	&	0.62 (0.11)	\\	\hline
            B.	&	B.	&	1.1 bn	&	0.04 (0.12)	&	0.20 (0.18)	&	0.77 (0.04)	&	0.59 (0.09)	\\	\hline
            D.	&	D.	&	66 mm	&	0.09 (0.22)	&	0.24 (0.25)	&	0.78 (0.06)	&	0.60 (0.11)	\\	\hline
            t5	&	t5-base	&	11 bn	&	0.06 (0.14)	&	0.23 (0.23)	&	0.80 (0.05)	&	0.63 (0.10)	\\	\hline
        \end{tabular}
    }
    
    \caption{\textbf Model performance scored on 100 observations sampled randomly from AdversarialQA(held-out test set) data. Model consists of BERT (named as B.), DistilBERT (named as D.), and T5. Tokenizer (named as token.) consists of bert-base-cased (named as B.), distilbert-base-uncased (named as D.), and t5-base. Param. is the model parameters of each model. Cosine is the cosine similarity, BERT is the STS (BERT) score, OpenAI is the STS (OpenAI) score, and Palm is the STS (Palm) score. The name ``bn'' and ``mm'' refer to billion and million, respectively.}
    \label{tab:test-models-on-adversarialQA-data}
\end{table}

Despite its CO2 emissions with the T4 GPU, the T5 model trained on the A100 GPU emitted less CO2 than models utilizing a BERT tokenizer and demonstrated a significantly better performance (see Table \ref{tab:model-a-tokenizer-b-test-set-performance-A100}). It outperformed BERT with a bert-base-cased tokenizer with a reduction in loss of 66.8\% and BERT with a distilbert-base-uncased tokenizer with a reduction in loss of 68\%.

Another remarkable result from our model training and testing was that using the A100 GPU had negligible impact on model performance, and in certain cases, models trained with the A100 GPU even outperformed those trained with the T4 GPU. However, for the majority of cases, the loss values for each model remained relatively consistent across both GPUs. The consistent performance of models between both GPUs when considered in combination with the lowered training time and carbon emissions of models trained on the A100 GPU evidences the possibility of lowering CO2 emissions without sacrificing model performance in LLM training.

\subsection{Discussion of CO2 Mitigation Strategies}

Although it may seem that based on the T5 model's low validation loss and high CO2 emissions, it is unrealistic to have a model be both optimized for performance and low CO2, this does not mean the ideal of achieving high performance while being environmentally conscientious is impossible. Based on the significant improvements on BERT brought by its distilled counterpart, DistilBERT, there may be potential to decrease CO2 emissions in the future through efforts to reduce the model parameters of existing models, thereby capitalizing on the performance benefits of high-performance models that currently produce significant CO2 emissions while simultaneously allowing for the mitigation of their environmental consequences.

This could be applied to various other models as well, including some of the other prominent LLMs that were not mentioned in the conducted training and testing of models in this paper. Transformer-XL, for example, enables learning dependency beyond a fixed length without disrupting temporal coherence. It achieves better performance on both short and long sequences and is up to 1,800+ times faster than vanilla Transformers during evaluation \cite{dai2019transformer}. Another model, Albert, makes use of two parameter-reduction techniques to lower memory consumption and increase the training speed of BERT \cite{lan2019albert}. These models may fit certain niches better than the models we tested and be fine-tuned for optimization on specific tasks. This is why we propose the implementation of certain strategies like lightening existing models and using faster GPUs that are generally applicable so that regardless of the model being used, corporations can stay environmentally conscious and have methods available for reducing the CO2 emissions of training their models. 

Additionally, the results achieved with the A100 GPU highlight the significant benefits of employing faster hardware for environmental sustainability and lowering training time; The A100 GPU proved to significantly increase CO2 emission efficiency, all without compromising the models' overall performance. Despite this, it is important to acknowledge a notable downside associated with utilizing an A100 GPU in contrast to a T4 GPU.

\subsection{Ethical Issues (Price)}
One prominent drawback of utilizing the A100 GPU for training is that the net cost of using it is higher. In our experiment, the A100 GPU increased RAM usage, which can result in higher expenses for companies and individuals utilizing it for model training. Additionally, the base price to buy the GPU is also more expensive, e.g. one listing on Amazon sells the GPU for \$6,799.\footnote{For the listing of the A100 GPU, please see \url{https://www.amazon.com/NVIDIA-Tesla-A100-Ampere-Graphics/dp/B0BGZJ27SL}}. This is comparatively much more expensive than the T4 GPU sold on the same website for a listing of \$1,175.99.\footnote{For the listing of the T4 GPU, please see \url{https://www.amazon.com/PNY-Datacenter-Express-Passive-Cooling/dp/B07QF9MJFR}.} This factor poses a concern, as the cost implications might limit its feasibility as an accessible option for many. This begs the question of whether training large language models in an environmentally sustainable way is truly an affordable option for any individual looking to conduct NLP model training. The trade-off between its environmental benefits and the associated financial implications necessitates careful consideration and further exploration of cost-effective alternatives to make sustainable AI practices more widely attainable.

With that said, for individuals without the budget to use a more expensive, faster GPU to decrease emissions, there are still other options. In fact, using a lighter model or modifying an existing model to have fewer model parameters are both viable strategies for saving on computational costs, as well as providing the added benefit of lowering CO2 emissions.

\section{Conclusion}

By implementing straightforward yet effective strategies, such as employing lighter models and faster GPUs, we can significantly lower a model's CO2 emissions without compromising model robustness and performance. To do so comes with certain drawbacks, such as a required budget needing to be allocated for better GPUs. It additionally requires the selection of lighter models or at least the expenditure of effort into reducing model parameters of existing models. However, this is a necessary step forward for the NLP field as embracing environmentally safe practices while maintaining results will not only be necessary to reduce the environmental damage of large language model training, but it also promises new horizons for sustainable and robust, high-performing models in the NLP field through added benefits in model efficiency and speed.

\bibliographystyle{abbrv}
\bibliography{ref}


\end{document}